\documentclass[letterpaper, 10 pt, conference]{ieeeconf}  
\IEEEoverridecommandlockouts                   
\usepackage{color,array}
\usepackage{amsmath, amssymb, paralist}
\usepackage{bm}
\usepackage{graphicx} 
\usepackage{comment}
\title{Structural-Ambiguity-Aware Translation from Natural Language to \\ Signal Temporal Logic}

\author{Kosei Fushimi, Kazunobu Serizawa, Junya Ikemoto, Kazumune Hashimoto\thanks{The authors are with the Graduate School of Engineering, The University of Osaka, Suita, Japan (Kazumune Hashimoto is the corresponding author: e-mail: hashimoto@eei.eng.osaka-u.ac.jp).}
}

\begin{document}

\maketitle
\thispagestyle{empty}
\pagestyle{empty}


\begin{abstract}
Signal Temporal Logic (STL) is widely used to specify timed and safety-critical tasks for cyber-physical systems, but writing STL formulas directly is difficult for non-expert users. Natural language (NL) provides a convenient interface, yet its inherent structural ambiguity makes one-to-one translation into STL unreliable. In this paper, we propose an \textit{ambiguity-preserving} method for translating NL task descriptions into STL candidate formulas. The key idea is to retain multiple plausible syntactic analyses instead of forcing a single interpretation at the parsing stage. To this end, we develop a three-stage pipeline based on Combinatory Categorial Grammar (CCG): ambiguity-preserving $n$-best parsing, STL-oriented template-based semantic composition, and canonicalization with score aggregation. The proposed method outputs a deduplicated set of STL candidates with plausibility scores, thereby explicitly representing multiple possible formal interpretations of an ambiguous instruction. In contrast to existing one-best NL-to-logic translation methods, the proposed approach is designed to preserve attachment and scope ambiguity. Case studies on representative task descriptions demonstrate that the method generates multiple STL candidates for genuinely ambiguous inputs while collapsing unambiguous or canonically equivalent derivations to a single STL formula.
\end{abstract}

\section{Introduction}
Autonomous robots, such as mobile robots, drones, and automated vehicles, are increasingly expected to execute complex tasks while satisfying timing and safety requirements. Formal specification languages provide a rigorous interface between high-level task descriptions and low-level control, and Signal Temporal Logic (STL) has become a standard choice for continuous-state systems because it can express time-bounded reachability, safety, sequencing, and persistence requirements over real-valued signals \cite{maler2004monitoring,donze2010robust}. In addition, STL is equipped with quantitative robustness semantics, which measure how strongly a trajectory satisfies a specification and therefore make STL attractive for motion planning, model predictive control, and controller synthesis \cite{Raman2014MPCSTL,Lin2023ModelPredictiveRobustness,Leung2023STLCG}.

Despite these advantages, writing STL formulas directly is difficult for non-expert users. One must choose atomic propositions, temporal operators, time intervals, and logical scopes precisely, even for seemingly simple task descriptions. Natural language (NL) provides an appealing front end for specifying tasks. Recent work in robotic language grounding and formal task specification has explored translating or grounding user instructions into temporal-logic formulas for downstream planning and execution \cite{lignos2015provably_correct_reactive_control,Wang2021GroundedLTLParser,liu2023lang2ltl,pan2023data_efficient_nl2ltl}. For a broader overview of robotic language grounding, see \cite{cohen2024robotic_language_grounding}. However, natural language is inherently ambiguous. In particular, a modifier or safety condition such as ``while avoiding A" may apply either to a single subtask or to the entire task, leading to multiple plausible formal interpretations of the same command. If this ambiguity is ignored, the resulting temporal-logic formula may fail to capture the intended control requirement.

Existing NL-to-TL/STL methods have mainly treated translation as a one-best prediction problem. This trend appears in learning-based translators based on neural models or large language models \cite{He2022DeepSTL,Chen2023NL2TL,Mao2024NL2STL,Fang2025KGST}, as well as in recent STL translation frameworks \cite{Mohammadinejad2025SystematicTranslationSTL}. This formulation is effective when the intended formalization is unique, but it becomes problematic for commands such as ``Within 10 seconds, reach B or reach C while avoiding A", whose modifier scope is not uniquely determined. Interactive approaches partly address this issue: Gavran \emph{et al.} \textit{nl2spec}  synthesize LTL specifications from natural language and demonstrations, while uses large language models to detect and resolve ambiguity through editable subtranslations \cite{Gavran2020InteractiveTemporalSpecs,Cosler2023nl2spec}. In the STL setting, Mohammadinejad \emph{et al.} consider interactive learning from language and demonstrations \cite{Mohammadinejad2022DialogueSTL}. A remaining challenge, however, is how to obtain a small and interpretable set of plausible formalizations when a single sentence admits multiple \textit{structurally} valid readings. In such cases, the desirable alternatives are those whose differences correspond to clear linguistic choices, such as modifier attachment or scope. Otherwise, it becomes difficult to distinguish genuine ambiguity in the input sentence from uncertainty introduced by the translation procedure.

In this paper, we address this gap by proposing a \textit{structural-ambiguity-aware} translation method from natural language to STL. The key idea is to preserve ambiguity rather than collapse it at the parsing stage. Specifically, we adapt Combinatory Categorial Grammar (CCG)-based semantic parsing (see, e.g., \cite{ccg2lambda}) to NL-to-STL translation by retaining multiple high-scoring parse trees, composing STL-oriented intermediate meanings with task-specific lexical-semantic templates, and canonicalizing equivalent derivations into a scored set of STL candidates. As a result, the output of the proposed method is not a single STL formula but a candidate set that explicitly represents multiple plausible formalizations of an ambiguous instruction.

This paper makes two main technical contributions. First, we propose an ambiguity-preserving formulation of NL-to-STL translation in which the output is a scored candidate set of STL formulas rather than a single formula. Second, we develop a symbolic pipeline that realizes this formulation by combining $n$-best CCG parsing, STL-oriented template-based semantic composition, and canonicalization with score aggregation. Representative case studies demonstrate that the proposed method produces multiple candidates for genuinely ambiguous inputs while collapsing unambiguous or canonically equivalent derivations to a single STL formula.

The remainder of this paper is organized as follows. Section~II reviews preliminaries on STL. Section~III introduces the problem setup and motivating ambiguity patterns. Section~IV presents the proposed ambiguity-preserving NL-to-STL translation pipeline. Section~V reports numerical evaluation on representative sentences. Finally, Section~VI concludes the paper and discusses limitations and future work.


\section{PRELIMINARIES}
\label{sec:prep}

\subsection{Signal Temporal Logic}
Signal Temporal Logic (STL) is a formal language for specifying temporal properties over real-valued signals. Since the main focus of this paper is ambiguity-preserving translation from natural-language control descriptions to STL, we briefly review only the STL syntax and robustness semantics used in the sequel. For standard definitions and notation of STL, the reader is referred to \cite{maler2004monitoring,donze2010robust}. STL formulas $\varphi$ are defined recursively from atomic propositions, logical operators, and temporal operators as
\begin{equation}
\begin{aligned}
\varphi ::= \ & \top \mid \mu \mid \neg \varphi \mid \varphi_1 \wedge \varphi_2 \mid \varphi_1 \vee \varphi_2 \\
& \mid \mathbf{F}_I \varphi \mid \mathbf{G}_I \varphi \mid \varphi_1 \mathbf{U}_I \varphi_2 ,
\end{aligned}
\label{eq:stl_syntax}
\end{equation}
where $\top$ denotes the Boolean constant \emph{True}, and $\neg$, $\wedge$, and $\vee$ denote negation, conjunction, and disjunction, respectively. The atomic proposition $\mu : \mathbb{R}^n \to \mathbb{B}$ is a Boolean-valued predicate over the current signal value. For example, given a real-valued function $h : \mathbb{R}^n \to \mathbb{R}$, we define
\begin{equation}
\mu(z)=\mathrm{True}\iff h(z)>0,\ z\in\mathbb{R}^n.
\label{eq:atomic_predicate}
\end{equation}

The operators $\mathbf{F}_I$, $\mathbf{G}_I$, and $\mathbf{U}_I$ denote the temporal operators \emph{eventually}, \emph{always}, and \emph{until}, respectively, over a discrete-time interval
\begin{equation}
I=[a,b]=\{\tau \in \mathbb{Z}_{\ge 0}\mid a \le \tau \le b\},
\ a,b \in \mathbb{Z}_{\ge 0}.
\label{eq:interval_def}
\end{equation}
In natural-language terms, $\mathbf{F}_I \varphi$ means that ``$\varphi$ holds at some time in $I$,'' $\mathbf{G}_I \varphi$ means that ``$\varphi$ holds at every time in $I$,'' and $\varphi_1 \mathbf{U}_I \varphi_2$ means that ``$\varphi_2$ holds at some time in $I$, while $\varphi_1$ holds up to that time.''
Moreover, the \emph{eventually} and \emph{always} operators can be expressed using the \emph{until} operator as
\begin{equation}
\mathbf{F}_I \varphi := \top \,\mathbf{U}_I\, \varphi,
\ 
\mathbf{G}_I \varphi := \neg \mathbf{F}_I \neg \varphi.
\label{eq:FG_def}
\end{equation}
The degree of satisfaction of an STL formula can be quantified by its robustness. A positive robustness value indicates satisfaction, whereas a negative value indicates violation. For a trajectory
\(
x = x_{0:T} = (x_0,x_1,\dots,x_T)
\)
and time $t$, the robustness $\rho^\varphi(x,t)$ is defined recursively as
\begin{align}
    & \rho^\mu(x, t) = h (x_t), \notag \\
    & \rho^{\neg \mu}(x, t) = -h (x_t),\notag \\ 
    & \rho^{\varphi_1 \land \varphi_2}(x, t) = \min (\rho^{\varphi_1} (x, t), \rho^{\varphi_2} (x, t)),\notag \\
    & \rho^{\varphi_1 \lor \varphi_2}(x, t) = \max (\rho^{\varphi_1}(x, t), \rho^{\varphi_2} (x, t)),\notag \\
    & \rho^{\bm{F}_{I}\varphi}(x, t) = \max_{t_1\in t+I}\rho^\varphi(x, t_1),\notag \\
    & \rho^{\bm{G}_{I}\varphi}(x, t) = \min_{t_1\in t+I}\rho^\varphi(x, t_1),\notag \\
    & \rho^{\varphi_1 \bm{U}_{I}\varphi_2}(x, t) = \max_{t_1\in t+I}\Bigl(\min (\rho^{\varphi_2}(x, t_1),  \min_{t_2 \in [t, t_1]}\rho^{\varphi_1}(x, t_2))\Bigr).\notag
\end{align}
with $t+I := \{\, t+\tau \mid \tau \in I \,\}$. 

\section{PROBLEM SETUP}
\label{sec:problem}

\subsection{Target Descriptions and Structural Ambiguity}

Translating natural-language (NL) control descriptions into STL is attractive because natural language is easy for users to provide, whereas writing STL formulas directly requires familiarity with temporal operators, logical scope, and task-dependent atomic propositions.

Natural language is inherently ambiguous. Among the different types of ambiguity, we focus on \emph{structural ambiguity}, i.e., ambiguity caused by attachment or scope differences in syntactic structure. This type of ambiguity is particularly important in NL-to-STL translation because STL is highly sensitive to scope and compositional structure.

In what follows, we use navigation/path-planning as a representative application domain. Specifically, we restrict attention to natural-language control descriptions built from task predicates such as \emph{reach} and \emph{avoid}, temporal modifiers such as \emph{within 10 seconds}, and connectives such as \emph{and}, \emph{or}, and \emph{while}. This class captures many navigation/path-planning instructions while keeping the problem setting well-defined.
Although the examples and numerical experiments are drawn from this domain, the formulation itself is not limited. Note that our focus is on preserving structurally distinct interpretations and translating them into STL candidates. The same idea can be applied to other control descriptions expressible in STL by changing the mapping from words to predicates and atomic propositions.

As a motivating example, consider the following task described in natural-language:
\begin{center}
\emph{Within 10 seconds, reach B or reach C while avoiding A.}
\end{center}
For this example, let $\phi_A$, $\phi_B$, and $\phi_C$ denote the atomic propositions that the agent is in regions $A$, $B$, and $C$, respectively. Then the sentence admits at least the following two interpretations, depending on the scope of the phrase \emph{while avoiding A}:
\begin{align}
\varphi_{1}
&= \mathbf{F}_{[0,10]}\phi_B \vee\
\bigl(\mathbf{F}_{[0,10]}\phi_C \wedge \mathbf{G}_{[0,10]}\neg\phi_A\bigr),
\label{eq:stl_local}
\\
\varphi_{2}
&= \mathbf{F}_{[0,10]}\bigl(\phi_B \vee \phi_C\bigr)\ \wedge\
\mathbf{G}_{[0,10]}\neg\phi_A.
\label{eq:stl_global}
\end{align}
The formula $\varphi_{1}$ corresponds to a \emph{local}-scope interpretation in which the avoidance condition is associated only with reaching $C$, whereas $\varphi_{2}$ corresponds to a \emph{global}-scope interpretation in which the avoidance condition applies to the entire disjunctive task. Although both interpretations are linguistically plausible, they impose different constraints on the system and may therefore lead to different planned trajectories. This example illustrates the central problem considered in this paper: a single natural-language description may correspond to multiple distinct yet plausible STL formulas.

\subsection{Problem Formulation}

Given an input natural-language description $d$, our goal is to generate a set of candidate STL specifications
\begin{equation}
\mathcal{C}(d)=\{(\varphi_j,s_j)\}_{j=1}^{m},
\label{eq:output_set}
\end{equation}
where $\varphi_j$ is an STL specification and $s_j \in \mathbb{R}_{\ge 0}$ is its plausibility score. Intuitively, the score $s_j$ indicates how strongly $\varphi_j$ is supported as an interpretation of $d$ by the translation pipeline. Notice that it is used to rank the candidates and does not measure control performance or STL robustness. When needed, these scores can be normalized into probabilities
\begin{equation}
p_j = \frac{s_j}{\sum_{k=1}^{m} s_k},
\qquad
\sum_{j=1}^{m} p_j = 1,
\label{eq:normalized_prob}
\end{equation}
so that $p_j$ represents the relative probability of $\varphi_j$ among the candidates generated.

\subsection{Practical Benefits}
\textit{Why is generating multiple candidates useful?} This is because ambiguity in language can lead to qualitatively different control behaviors. In the motivating example, a planner using $\varphi_{1}$ may accept a trajectory that reaches $B$ even if it passes through $A$, whereas a planner using $\varphi_{2}$ must reject any trajectory that visits $A$. Thus, if the system prematurely commits to a single STL formula, it may produce a behavior that is formally correct for that formula but inconsistent with the user's true intent.

By generating multiple STL candidates, the system can present not only the formulas themselves but also \textit{candidate trajectories} derived from them. This allows the user to compare possible behaviors, identify unintended interpretations before execution, and select the desired specification. In this sense, the candidate set serves as an interpretable interface between ambiguous NL task description and formal control synthesis.

\section{Ambiguity-Preserving NL-to-STL Translation}
\label{sec:method}

\begin{figure}[t]
  \centering
  \includegraphics[width=0.99\linewidth]{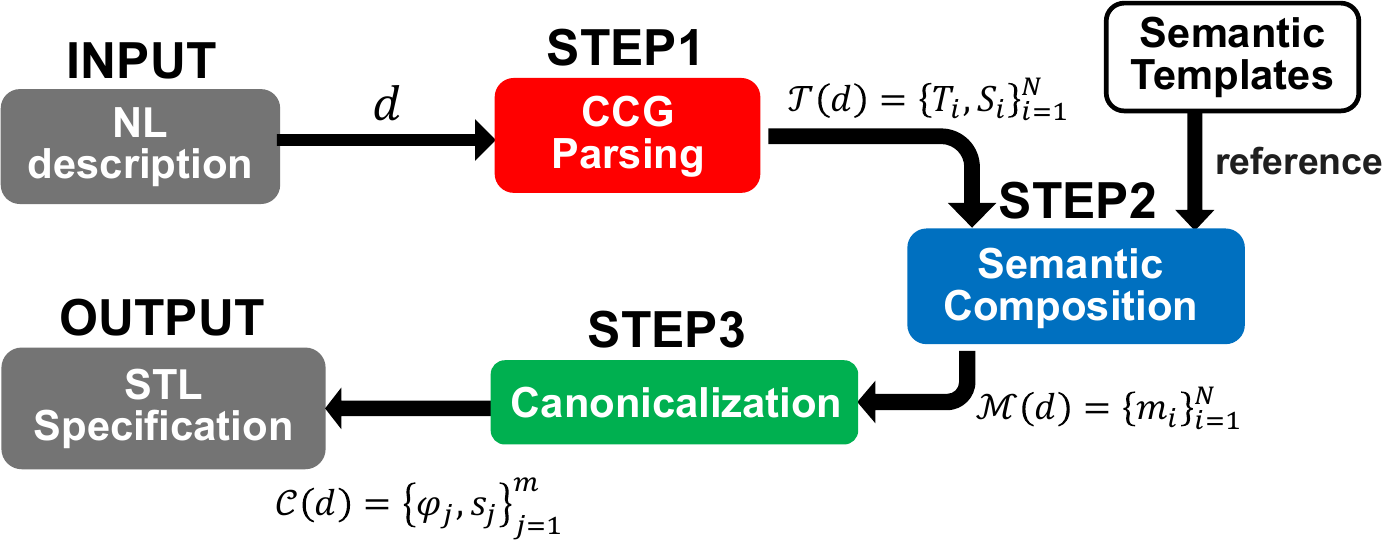}
  \caption{Overview of the Proposed Method.}
  \label{fig:pipeline}
\end{figure}
\subsection{Overview}
In this paper, we leverage syntactic parsing \cite{ccg2lambda}, which is a standard technique in natural language processing, to capture structural ambiguity and preserve multiple plausible interpretations of an input sentence. An overview of the proposed method is shown in Fig.~\ref{fig:pipeline}. Given an input natural-language description $d$, the method generates STL candidates through a three-stage pipeline: first, it constructs a set $\mathcal{T}(d)$ of plausible parse trees and the associated scores (Step~1); next, it derives a set $\mathcal{M}(d)$ of STL-oriented meaning representations through semantic composition (Step~2); and finally, it converts them into a deduplicated set $\mathcal{C}(d)$ of STL candidates with plausibility scores (Step~3).

Although the basic ingredients build on standard CCG-based semantic parsing, the proposed method is not a direct application of an off-the-shelf CCG parser to obtain a single syntactic analysis, in the sense that we adapt CCG-based parsing for ambiguity-preserving NL-to-STL translation. Specifically, as described in the following subsections, the proposed pipeline (i) retains multiple high-scoring parse trees so that attachment and scope ambiguities are preserved, (ii) composes STL-oriented meaning representations using task-specific semantic templates, and (iii) canonicalizes semantically equivalent derivations and aggregates their scores into a deduplicated set of STL candidates. In this way, syntactically distinct readings of the same sentence can be propagated to distinct STL specifications when appropriate, while spurious duplication is suppressed when different derivations yield the same STL formula. Each stage of the process is described in detail below.


\subsection{Ambiguity-Preserving $n$-best CCG Parsing}
In Step~1, the input sentence $d$ is parsed by a probabilistic CCG parser, and multiple high-scoring parse trees are retained rather than committing to a single best analysis. We use Combinatory Categorial Grammar (CCG) \cite{ccg2lambda} as a structured parsing backend. In this study, the role of CCG is not to provide a detailed linguistic analysis, but to explicitly represent attachment and scope differences in a form that can be propagated to STL generation.
A CCG parse tree records how words and phrases combine to form a sentence. When a sentence is structurally ambiguous, the parser can return multiple plausible parse trees based on the parsing scores rather than a single one. Given an input sentence $d$, we obtain an $n$-best set of parse trees
\begin{equation}
\mathcal{T}(d)=\{T_i, S_i\}_{i=1}^{N},
\label{eq:parse_set}
\end{equation}
where each $T_i$ represents one plausible syntactic interpretation of $d$, and $S_i \in \mathbb{R}$ denotes the parsing score associated with $T_i$. In our implementation using depccg \cite{depccg}, $S_i$ is the parser-internal score assigned by the underlying locally factored CCG model, which combines token-level supertag scores and dependency-head scores to rank candidate parses.

As an example, let us consider again the following sentence:
\begin{center}
\emph{Within 10 seconds, reach B or reach C while avoiding A.}
\end{center}
In this sentence, the phrase \emph{while avoiding A} may attach either to the subexpression \emph{reach C} or to the entire phrase \emph{reach B or reach C}. These two attachments correspond to different parse trees. Since the proposed method retains multiple parse trees instead of selecting only the top one, both interpretations can be preserved and passed to the semantic composition stage.

\subsection{STL-oriented template-based semantic composition}
In Step~2, an STL-oriented meaning representation is constructed for each parse tree by semantic composition based on task-specific lexical-semantic templates \cite{ccg2lambda}. Given each parse tree $T_i \in \mathcal{T}(d)$, we construct an STL-oriented meaning representation by semantic composition. In CCG-based semantic parsing, semantic composition assigns a lexical-semantic template to each lexical item or phrase and composes these templates according to the combinatory structure of the derivation. Here, a semantic template is a rule that maps a lexical item (together with its CCG category) to an STL-oriented lambda expression. 

Basic task expressions are mapped onto intermediate temporal operators such as $\mathrm{F}(I,X)$ and $\mathrm{G}(I,X)$, whereas functional expressions are mapped onto lambda abstractions such as $\lambda Q.\lambda P.\,(P \wedge Q)$, $\lambda Q.\lambda P.\,(P \vee Q)$, $\lambda Q.\lambda P.\,\mathrm{SEQ}(P,Q)$, and $\lambda P.\,P(I(a,b))$. Here, we use standard lambda-calculus notation for compositional semantics: $\lambda x.\,f(x)$ denotes the function that maps an argument $x$ to $f(x)$, and $\lambda Q.\lambda P.\,f(P,Q)$ denotes a higher-order function that successively takes two semantic arguments $Q$ and $P$. In this notation, $P$ and $Q$ denote intermediate semantic representations of subtasks, $\mathrm{SEQ}(P,Q)$ denotes sequential composition, and $P(I(a,b))$ denotes the application of the time interval $I(a,b)$ to the temporal operator $P$. Moreover, $P \wedge Q$ and $P \vee Q$ denote logical conjunction and logical disjunction, respectively. These operators are intermediate semantic constructs used in Step~2 and are translated into final STL formulas in Step~3.

In our implementation, semantic composition is performed using \texttt{ccg2lambda}~\cite{ccg2lambda}. While \texttt{ccg2lambda} is originally designed to derive linguistically rich semantic representations, we design task-specific semantic templates tailored to NL-to-STL translation. Applying these templates to each parse tree yields
\begin{equation}
\mathcal{M}(d)=\{m_i\}_{i=1}^{N},
\label{eq:semantic_set}
\end{equation}
where each $m_i$ is an STL-oriented meaning representation obtained by composing the lexical-semantic templates along the derivation $T_i$. Importantly, $m_i$ is an intermediate representation and need not yet be a complete well-formed STL formula; final conversion, deduplication, and score aggregation are deferred to Step~3. Each $m_i$ also inherits the support score of its source derivation, which is later aggregated at the formula level.

\begin{table}[t]
\centering
\small
\caption{Representative STL-oriented semantic templates. The symbol $\neg$ denotes the negation operator.}
\label{tab:semantic_roles}
\begin{tabular}{p{2.6cm}p{4.1cm}}
\hline
Expression & Abstract template \\
\hline
\emph{reach X} & $\lambda I.\mathrm{F}(I,X)$ \\
\emph{avoid X} & $\lambda I.\mathrm{G}(I,\neg X)$ \\
\emph{or} & $\lambda Q.\lambda P.(P \vee Q)$ \\
\emph{and then} & $\lambda Q.\lambda P.\mathrm{SEQ}(P,Q)$ \\
\emph{while} & $\lambda Q.\lambda P.(P \wedge Q)$ \\
\emph{within $T$ seconds} & $\lambda P.P(I(0,T))$ \\
\hline
\end{tabular}
\end{table}

Table~\ref{tab:semantic_roles} lists representative semantic templates used in our implementation. The table is illustrative rather than exhaustive and highlights the template patterns that determine task structure, scope, and temporal constraints in NL-to-STL translation. A key point is that the templates themselves are fixed; ambiguity arises from how they are composed along the parse tree.

As a result, different parse trees give rise to different scopes of operators such as $\wedge$, $\vee$, and $\mathrm{SEQ}$, thereby yielding different intermediate semantic representations.

In the motivating example, one parse associates the modifier \emph{while avoiding A} only with \emph{reach C}, whereas another associates it with the entire disjunctive phrase \emph{reach B or reach C}. Thus, even though the same abstract templates in Table~\ref{tab:semantic_roles} are used in both cases, their different combinatory structures lead to different elements of $\mathcal{M}(d)$, which are then converted into different STL candidates in Step~3. In \texttt{ccg2lambda}, the actual templates are implemented as YAML-based rules; here, we abstract away from those implementation details and focus on the representative abstract template patterns summarized in Table~\ref{tab:semantic_roles}.

\subsection{Conversion to STL, Canonicalization, and Score Aggregation}

In Step~3, the intermediate meaning representations in $\mathcal{M}(d)$ are converted into STL formulas, canonicalized, and scored at the formula level. In particular, the intermediate operators introduced in Step~2, such as $\mathsf{F}$, $\mathsf{G}$, and $\mathsf{SEQ}$, are resolved into standard STL syntax.
Because the elements of $\mathcal{M}(d)$ are STL-oriented intermediate representations rather than final STL formulas, some $m_i \in \mathcal{M}(d)$ may be incomplete or ill-formed from the viewpoint of STL syntax, and distinct derivations may still yield the same STL formula after conversion. Step~3 resolves both issues through conversion, canonicalization, and score aggregation.

For each $m_i$, we first attempt to convert it into a complete STL formula. Representations that cannot be converted into a well-formed STL formula are discarded. The remaining formulas are then canonicalized so that derivationally different outputs that are equivalent under our canonicalization rules are mapped to a common normal form. This prevents the final candidate set from being artificially inflated by duplicate formulas arising from different parse trees or composition paths.

Finally, scores are aggregated at the formula level. Because each $m_i$ inherits the support score of its source derivation, multiple derivations that collapse to the same canonical STL formula contribute jointly to a single candidate score. As a result, the final set in Eq.~\eqref{eq:output_set} contains only unique STL formulas together with plausibility scores that reflect their total support from the ambiguity-preserving translation pipeline. In this way, Step~3 preserves meaningful ambiguity while suppressing spurious duplication.

\section{Numerical evaluation on Representative Sentences}

\subsection{Setting}
To evaluate the proposed ambiguity-preserving NL-to-STL translation pipeline, we consider representative natural-language task descriptions. 
Specifically, the sentence set covers simple reachability, time-bounded tasks, disjunction, sequential composition, and modifier attachment by \emph{while avoiding A}. The aim of this section is not to provide a large-scale benchmark, but to verify three properties of the proposed method: (i) it produces multiple STL candidates when the input contains genuine structural ambiguity, (ii) it yields a single canonical STL candidate when the input is effectively unambiguous at the STL level, and (iii) it merges derivationally different outputs when they reduce to the same canonical STL formula.

For CCG parsing, we use \texttt{depccg} \cite{depccg} and retain up to the top $N=40$ parse trees for each input sentence, yielding a set $\mathcal{T}(d)$ as in Eq.~\eqref{eq:parse_set}. Each parse tree is then mapped to an STL-oriented meaning representation and processed by the canonicalization and score-aggregation procedure described in Step~3. Each final candidate $\varphi \in \mathcal{C}(d)$ is assigned an aggregated score obtained by summing the support of all derivations that reduce to $\varphi$, and the aggregated scores are normalized according to Eq.~\eqref{eq:normalized_prob} to obtain $p(\varphi \mid d)$.

The 12 representative input sentences are as follows:
\begin{enumerate}
  \item[\textbf{S1.}] Reach B within 10 seconds.
  \item[\textbf{S2.}] Reach B within 10 seconds while avoiding A.
  \item[\textbf{S3.}] Within 10 seconds, reach B while avoiding A.
  \item[\textbf{S4.}] Reach  B or C within 10 seconds.
  \item[\textbf{S5.}]Reach B within 10 seconds or reach C within 15 seconds.
  \item[\textbf{S6.}] Reach B within 10 seconds and then reach C within 15 seconds.
  \item[\textbf{S7.}] Reach B within 10 seconds and then reach C within 15 seconds and then reach D within 5 seconds.
  \item[\textbf{S8.}] Within 10 seconds, reach B or reach C while avoiding A.
  \item[\textbf{S9.}] Reach B within 10 seconds or reach C within 15 seconds while avoiding A.
  \item[\textbf{S10.}]Reach B within 10 seconds and then reach C within 15 seconds while avoiding A.
  \item[\textbf{S11.}] Reach B within 10 seconds and then reach C within 15 seconds and then reach D within 5 seconds while avoiding A.
  \item[\textbf{S12.}] Reach B within 10 seconds and then reach C within 15 seconds or reach D within 5 seconds while avoiding A.
\end{enumerate}

Among them, S8, S9, S10, S11, S12 correspond to the two motivating ambiguity patterns considered in this paper, i.e., modifier scope over a disjunctive task description and modifier scope over a sequential task description, respectively.

\subsection{Summary of Results}
Table~\ref{tab:representative_eval} summarizes the outputs obtained for the 12 representative sentences. For each sentence, the column ``Type'' indicates the primary syntactic or semantic phenomenon targeted by that sentence, such as simple reachability, modifier attachment, disjunction, sequential composition, or scope ambiguity. We also report the number of manually identified reference interpretations (\#Ref.), the number of unique STL candidates after canonicalization ($|\mathcal{C}(d)|$), and whether the reference interpretation set is recovered. This summary is intended to show not only that the proposed method can generate multiple STL formulas when needed, but also that it avoids unnecessary branching when the input is effectively unambiguous at the STL level.

\begin{table}[t]
\centering
\small
\caption{Summary of results on representative sentences.}
\label{tab:representative_eval}
\setlength{\tabcolsep}{4pt}
\begin{tabular}{c p{3cm} c c c}
\hline
ID & Type & \#Ref. & $|\mathcal{C}(d)|$ & Recovered \\
\hline
S1  & reachability(base)        & 1 & 1 & Yes \\
S2  & conjunction + modification (pattern 1)       & 1 & 1 & Yes \\
S3  & conjunction + modification (pattern 2) & 1 & 1 & Yes \\
S4  & same time + disjunction         & 1 & 1 & Yes \\
S5  & different time + disjunction          & 1 & 1 & Yes \\
S6  & sequence     & 1 & 1   & Yes \\
S7  & long sequence     & 1 & 1   & Yes \\
S8  & same time + disjunction (scope ambiguity)     & 2 & 2 & Yes \\
S9  & different time + disjunction (scope ambiguity)    & 2 & 2 & Yes \\
S10 & sequence (scope ambiguity)        & 2 & 2 & Yes \\
S11 & long sequence (scope ambiguity)      & 3 & 3 & Yes \\
S12 & sequence + disjunction (scope ambiguity)     & 5 & 5 & Yes \\
\hline
\end{tabular}
\end{table}

The key question is whether the proposed pipeline propagates ambiguity selectively. In particular, sentences with genuine attachment or scope ambiguity should yield multiple candidates, whereas structurally simpler sentences should collapse to a single canonical STL formula after Step~3. Such a result would indicate that the proposed method is not limited to the two motivating examples, but can be applied to a broader set of representative task descriptions while controlling over-generation through canonicalization and score aggregation. 
As a result, for all of these examples, the reference interpretation sets were successfully recovered.

\subsection{Detailed Outputs for Representative Ambiguous Cases}
Table~\ref{tab:ambiguous_results} reports the full STL formulas and normalized probabilities for S8, S9, S10, S11, S12. These 5 cases illustrate the target ambiguity patterns of this paper: modifier attachment to a disjunctive task description and modifier attachment to a sequential task description.
\begin{table*}[t]
\centering
\small
\caption{Generated STL formulas and normalized probabilities for representative ambiguous cases.}
\label{tab:ambiguous_results}
\setlength{\tabcolsep}{3pt}
\begin{tabular}{@{}p{0.65\textwidth}r@{}}
\hline
STL formula $\varphi$ & $p(\varphi \mid d)$ \\
\hline
\multicolumn{2}{@{}l@{}}{\textbf{S8:} \emph{Within 10 seconds, reach B or reach C while avoiding A.}} \\
$\mathbf{F}_{[0,10]}\phi_B \vee \bigl(\mathbf{F}_{[0,10]}\phi_C \wedge \mathbf{G}_{[0,10]}\neg\phi_A\bigr)$ & 0.6774565705 \\
$\mathbf{F}_{[0,10]}\bigl(\phi_B \vee \phi_C\bigr)\wedge \mathbf{G}_{[0,10]}\neg\phi_A$ & 0.3225434295 \\
\hline
\multicolumn{2}{@{}l@{}}{\textbf{S9:} \emph{Reach B within 10 seconds or reach C within 15 seconds while avoiding A.}} \\
$\mathbf{F}_{[0,10]}\phi_B \vee \bigl(\mathbf{F}_{[0,15]}\phi_C \wedge \mathbf{G}_{[0,15]}\neg\phi_A\bigr)$ & 0.86272391 \\
$\bigl(\mathbf{F}_{[0,10]}\phi_B \vee \mathbf{F}_{[0,15]}\phi_C\bigr)\wedge \mathbf{G}_{[0,15]}\neg\phi_A$ & 0.13727609 \\
\hline
\multicolumn{2}{@{}l@{}}{\textbf{S10:} \emph{Reach B within 10 seconds and then reach C within 15 seconds while avoiding A.}} \\
$\mathbf{F}_{[0,10]}\bigl(\phi_B \wedge \mathbf{F}_{[0,15]}\phi_C \wedge \mathbf{G}_{[0,15]}\neg\phi_A\bigr)$ & 0.9896166506 \\
$\mathbf{F}_{[0,10]}\bigl(\phi_B \wedge \mathbf{F}_{[0,15]}\phi_C\bigr)\wedge \mathbf{G}_{[0,25]}\neg\phi_A$ & 0.0103833494 \\
\hline
\multicolumn{2}{@{}l@{}}{\textbf{S11:} \emph{Reach B within 10 seconds and then reach C within 15 seconds and then reach D within 5 seconds while avoiding A.}} \\
$\mathbf{F}_{[0,10]}\bigl(\phi_B \wedge \mathbf{F}_{[0,15]}\bigl(\phi_C \wedge \mathbf{F}_{[0,5]}\phi_D \wedge\mathbf{G}_{[0,5]}\neg\phi_A\bigr)$ & 0.934112393 \\
$\mathbf{F}_{[0,10]}\bigl(\phi_B \wedge \mathbf{F}_{[0,15]}\bigl(\phi_C \wedge \mathbf{F}_{[0,5]}\phi_D\bigr) \wedge\mathbf{G}_{[0,20]}\neg\phi_A\bigr)$ & 0.0599828964 \\
$\mathbf{F}_{[0,10]}\bigl(\phi_B \wedge \mathbf{F}_{[0,15]}\bigl(\phi_C \wedge \mathbf{F}_{[0,5]}\phi_D\bigr) \bigr)\wedge\mathbf{G}_{[0,30]}\neg\phi_A$ & 0.0059047106 \\
\hline
\multicolumn{2}{@{}l@{}}{\textbf{S12:} \emph{Reach B within 10 seconds and then reach C within 15 seconds or reach D within 5 seconds while avoiding A.}} \\
$\mathbf{F}_{[0,10]}\bigl(\phi_B \wedge \bigl(\bigl(\mathbf{F}_{[0,15]}\phi_C \vee \mathbf{F}_{[0,5]}\phi_D\bigr)\wedge \mathbf{G}_{[0,15]}\neg\phi_A\bigr)\bigr)$ & 0.5004023346 \\
$\mathbf{F}_{[0,10]}\bigl(\phi_B \wedge \bigl(\mathbf{F}_{[0,15]}\phi_C \vee \bigl(\mathbf{F}_{[0,5]}\phi_D\wedge \mathbf{G}_{[0,5]}\neg\phi_A\bigr)\bigr)\bigr)$ & 0.478979327 \\
$\mathbf{F}_{[0,10]}\bigl(\phi_B \wedge \bigl(\mathbf{F}_{[0,15]}\phi_C \vee \mathbf{F}_{[0,5]}\phi_D\bigr)\bigr)\wedge \mathbf{G}_{[0,25]}\neg\phi_A$ & 0.015082217 \\
$\mathbf{F}_{[0,10]}\bigl(\phi_B \wedge \mathbf{F}_{[0,15]}\phi_C\bigr) \vee \bigl(\mathbf{F}_{[0,5]}\phi_D\wedge \mathbf{G}_{[0,5]}\neg\phi_A\bigr)$  & 	0.0054311103 \\
$\bigl(\mathbf{F}_{[0,10]}\bigl(\phi_B \wedge \mathbf{F}_{[0,15]}\phi_C\bigr) \vee \mathbf{F}_{[0,5]}\phi_D\bigr)\wedge \mathbf{G}_{[0,25]}\neg\phi_A$  & 	0.000105011 \\
\hline
\end{tabular}
\end{table*}

For all representative ambiguous examples, the generated candidate sets contain the reference interpretations identified manually. Although the resulting probabilities are not uniform, reflecting the preferences of the underlying parser, the very existence of multiple candidates indicates that the proposed method preserves structurally induced ambiguity up to the level of formal STL specifications. More broadly, this evaluation protocol is intended to verify that the same pipeline can be applied to representative sentence patterns beyond those considered here, while the canonicalization in Step~3 suppresses unnecessary duplication.
\subsection{Comparison with a ChatGPT-Based Baseline}
As a point of comparison, we evaluated a prompted general-purpose LLM baseline using ChatGPT on the five structurally ambiguous inputs S8--S12. The aim of this experiment is not to provide a comprehensive ranking over all learning-based NL-to-STL methods, but to examine whether a prompted LLM can enumerate the structurally licensed STL readings identified for these examples. The exact prompt and evaluation protocol are provided in Appendix~\ref{app:chatgpt_prompt}. Table~\ref{tab:chatgpt_summary} summarizes the sentence-level results, and Table~\ref{tab:stl_candidates} lists the ChatGPT outputs.
A returned formula was counted as \emph{correct} if and only if it matched one of the manually identified reference interpretations used for the proposed method. In the ChatGPT outputs, the remaining incorrect cases fall into two categories: (E1) \emph{scope/anchoring mismatch}, where the avoidance condition is attached at the wrong temporal level or with the wrong effective interval; and (E2) \emph{structurally unsupported extra reading}, where the output does not correspond to any reference parse-based interpretation.

\begin{table}[t]
\centering
\small
\caption{Sentence-level summary for the ChatGPT baseline on S8--S12. ``Recovered" denotes how many reference interpretations were recovered at least once among the returned formulas.}
\label{tab:chatgpt_summary}
\setlength{\tabcolsep}{4pt}
\begin{tabular}{lccccc}
\hline
ID & \#Out & Recovered & Correct & E1 & E2 \\
\hline
S8  & 2 & 2/2 & 2 & 0 & 0 \\
S9  & 2 & 2/2 & 2 & 0 & 0 \\
S10 & 2 & 0/2 & 0 & 1 & 1 \\
S11 & 3 & 3/3 & 3 & 0 & 0 \\
S12 & 5 & 4/5 & 4 & 1 & 0 \\
\hline
Total & 14 & 11/14 & 11 & 2 & 1 \\
\hline
\end{tabular}
\end{table}

The ChatGPT outputs recover all reference readings for S8, S9, and S11, but not for S10 and S12. In particular, S10 recovers none of the two reference readings, and S12 recovers four of the five reference readings. Overall, 11 of the 14 returned formulas match a reference interpretation after canonicalization, covering 11 of the 14 reference readings across S8--S12.
The main failure mode is incorrect temporal anchoring. For example, in S10 the formula
\[
\mathbf{F}_{[0,10]}
\bigl(
\phi_B \wedge
\mathbf{F}_{[0,15]}
(
\phi_C \wedge \mathbf{G}_{[0,15]}\neg\phi_A
)
\bigr)
\]
places $\mathbf{G}_{[0,15]}\neg\phi_A$ inside the inner eventuality, so the avoidance interval starts at the time of reaching $C$ rather than at the beginning of the intended second-step subtask. In S12, the formula
\[
\mathbf{G}_{[0,15]}\neg\phi_A
\wedge
\bigl(
\mathbf{F}_{[0,10]}(\phi_B \wedge \mathbf{F}_{[0,15]}\phi_C)
\vee
\mathbf{F}_{[0,5]}\phi_D
\bigr)
\]
does not match any reference interpretation either: it is too short for the top-level global reading, which requires $\mathbf{G}_{[0,25]}\neg\phi_A$, and it is placed outside the outer $\mathbf{F}_{[0,10]}$ unlike the post-$B$ disjunction-wide reading. The remaining unsupported case in S10 combines a first-step avoidance condition with the same mis-anchored second-step reading, yielding a composite formula not licensed by any reference parse.

These observations suggest that, although a prompted LLM can recover certain valid readings, it does not yet reliably enumerate exactly the structurally licensed STL candidates or recover them exhaustively under a fixed prompting protocol. Note that LLM outputs can vary substantially with prompt design, so the baseline results should be interpreted as prompt-dependent rather than as an intrinsic limit of ChatGPT. While additional prompt engineering may improve some cases, such sensitivity itself raises concerns about reproducibility and stable coverage of structurally licensed interpretations. In contrast, the proposed symbolic pipeline generates candidates only from explicitly represented parse-structural alternatives and recovers the complete reference set for these cases.

\begin{table*}[t]
\centering
\scriptsize
\caption{STL specifications generated by ChatGPT for representative ambiguous cases. The raw outputs are normalized to the manuscript notation $(\phi_A,\phi_B,\phi_C,\phi_D)$. ``Correct'' means that the output matches one of the manually identified reference interpretations after canonicalization.}
\label{tab:stl_candidates}
\setlength{\tabcolsep}{3pt}
\renewcommand{\arraystretch}{1.03}
\begin{tabular}{@{}p{0.67\textwidth}p{0.09\textwidth}p{0.19\textwidth}@{}}
\hline
STL formula $\varphi$ & Assessment & Note \\
\hline

\multicolumn{3}{@{}l@{}}{\textbf{S8:} \emph{Within 10 seconds, reach B or reach C while avoiding A.}} \\
$\mathbf{F}_{[0,10]}\phi_B \,\vee\, \bigl(\mathbf{G}_{[0,10]}\neg\phi_A \,\wedge\, \mathbf{F}_{[0,10]}\phi_C\bigr)$
& \textbf{Correct} & Matches the local-scope reference. \\
$\mathbf{G}_{[0,10]}\neg\phi_A \,\wedge\, \mathbf{F}_{[0,10]}(\phi_B \vee \phi_C)$
& \textbf{Correct} & Equivalent to the global-scope reference after canonicalization. \\
\hline

\multicolumn{3}{@{}l@{}}{\textbf{S9:} \emph{Reach B within 10 seconds or reach C within 15 seconds while avoiding A.}} \\
$\mathbf{F}_{[0,10]}\phi_B \,\vee\, \bigl(\mathbf{G}_{[0,15]}\neg\phi_A \,\wedge\, \mathbf{F}_{[0,15]}\phi_C\bigr)$
& \textbf{Correct} & Matches the local-scope reference. \\
$\mathbf{G}_{[0,15]}\neg\phi_A \,\wedge\, \bigl(\mathbf{F}_{[0,10]}\phi_B \,\vee\, \mathbf{F}_{[0,15]}\phi_C\bigr)$
& \textbf{Correct} & Matches the global-scope reference. \\
\hline

\multicolumn{3}{@{}l@{}}{\textbf{S10:} \emph{Reach B within 10 seconds and then reach C within 15 seconds while avoiding A.}} \\
$\mathbf{F}_{[0,10]}
\bigl(
\phi_B \,\wedge\,
\mathbf{F}_{[0,15]}
(
\phi_C \,\wedge\, \mathbf{G}_{[0,15]}\neg\phi_A
)
\bigr)$
& \textbf{E1} & The avoidance interval is anchored at the $C$-reaching instant, not at the beginning of the second-step subtask. \\
$\bigl(\mathbf{G}_{[0,10]}\neg\phi_A \,\wedge\, \mathbf{F}_{[0,10]}\phi_B\bigr)
\,\wedge\,
\mathbf{F}_{[0,10]}
\bigl(
\phi_B \,\wedge\,
\mathbf{F}_{[0,15]}
(
\phi_C \,\wedge\, \mathbf{G}_{[0,15]}\neg\phi_A
)
\bigr)$
& \textbf{E2} & Unsupported composite reading; matches neither the local-scope nor the global-scope reference. \\
\hline

\multicolumn{3}{@{}l@{}}{\textbf{S11:} \emph{Reach B within 10 seconds and then reach C within 15 seconds and then reach D within 5 seconds while avoiding A.}} \\
$\mathbf{F}_{[0,10]}
\bigl(
\phi_B \,\wedge\,
\mathbf{F}_{[0,15]}
\bigl(
\phi_C \,\wedge\,
(
\mathbf{G}_{[0,5]}\neg\phi_A \,\wedge\, \mathbf{F}_{[0,5]}\phi_D
)
\bigr)
\bigr)$
& \textbf{Correct} & Matches the local-$D$-scope reference. \\
$\mathbf{F}_{[0,10]}
\bigl(
\phi_B \,\wedge\,
(
\mathbf{G}_{[0,20]}\neg\phi_A
\,\wedge\,
\mathbf{F}_{[0,15]}
(
\phi_C \,\wedge\, \mathbf{F}_{[0,5]}\phi_D
)
)
\bigr)$
& \textbf{Correct} & Matches the $C\!\rightarrow\!D$ subsequence-scope reference. \\
$\mathbf{G}_{[0,30]}\neg\phi_A
\,\wedge\,
\mathbf{F}_{[0,10]}
\bigl(
\phi_B \,\wedge\,
\mathbf{F}_{[0,15]}
\bigl(
\phi_C \,\wedge\, \mathbf{F}_{[0,5]}\phi_D
\bigr)
\bigr)$
& \textbf{Correct} & Matches the global-scope reference. \\
\hline

\multicolumn{3}{@{}l@{}}{\textbf{S12:} \emph{Reach B within 10 seconds and then reach C within 15 seconds or reach D within 5 seconds while avoiding A.}} \\
$\mathbf{F}_{[0,10]}
\bigl(
\phi_B \,\wedge\,
(
\mathbf{F}_{[0,15]}\phi_C
\,\vee\,
(
\mathbf{G}_{[0,5]}\neg\phi_A \,\wedge\, \mathbf{F}_{[0,5]}\phi_D
)
)
\bigr)$
& \textbf{Correct} & Matches the post-$B$ local-$D$ reference. \\
$\mathbf{F}_{[0,10]}
\bigl(
\phi_B \,\wedge\, \mathbf{F}_{[0,15]}\phi_C
\bigr)
\,\vee\,
\bigl(
\mathbf{G}_{[0,5]}\neg\phi_A \,\wedge\, \mathbf{F}_{[0,5]}\phi_D
\bigr)$
& \textbf{Correct} & Matches the top-level local-$D$ reference. \\
$\mathbf{F}_{[0,10]}
\bigl(
\phi_B \,\wedge\,
(
\mathbf{G}_{[0,15]}\neg\phi_A
\,\wedge\,
(
\mathbf{F}_{[0,15]}\phi_C
\,\vee\,
\mathbf{F}_{[0,5]}\phi_D
)
)
\bigr)$
& \textbf{Correct} & Matches the post-$B$ disjunction-wide reference. \\
$\mathbf{G}_{[0,25]}\neg\phi_A
\,\wedge\,
\mathbf{F}_{[0,10]}
\bigl(
\phi_B \,\wedge\,
(
\mathbf{F}_{[0,15]}\phi_C
\,\vee\,
\mathbf{F}_{[0,5]}\phi_D
)
\bigr)$
& \textbf{Correct} & Matches the post-$B$ global-scope reference. \\
$\mathbf{G}_{[0,15]}\neg\phi_A
\,\wedge\,
\bigl(
\mathbf{F}_{[0,10]}
(
\phi_B \,\wedge\, \mathbf{F}_{[0,15]}\phi_C
)
\,\vee\,
\mathbf{F}_{[0,5]}\phi_D
\bigr)$
& \textbf{E1} & Matches neither reference: it is too short for the top-level global reading and placed at the wrong level for the post-$B$ disjunction-wide reading. \\
\hline

\multicolumn{3}{@{}p{\textwidth}@{}}{\footnotesize
\textbf{Assessment labels.}
\textbf{Correct}: matches a manually identified reference interpretation after canonicalization.
\textbf{E1}: scope/anchoring mismatch, i.e., the avoidance interval is attached at the wrong temporal level or with the wrong effective interval.
\textbf{E2}: structurally unsupported extra reading, i.e., the output does not correspond to any reference parse-based interpretation.} \\
\hline
\end{tabular}
\end{table*}

\section{Current limitations and future work}
We proposed an ambiguity-preserving symbolic method for translating natural-language task descriptions into STL formulas. The proposed pipeline combines $n$-best CCG parsing, STL-oriented template-based semantic composition, and canonicalization with score aggregation so as to generate a scored set of candidate STL formulas rather than committing to a single interpretation. In this way, structurally distinct readings of an ambiguous sentence can be propagated to distinct formal specifications while canonically equivalent derivations are merged into a compact candidate set. 

{Several limitations remain.} First, the current formulation assumes that the input sentence provides explicit temporal information that can be mapped to STL time bounds. As a result, task descriptions without explicit timing expressions require additional assumptions and are not yet handled in a principled manner in the present implementation. Second, the current template set covers only a limited class of task expressions and ambiguity patterns, mainly those involving reachability, avoidance, disjunction, sequential composition, and modifier scope. Extending the method to richer linguistic phenomena, such as implicit temporal constraints, context-dependent expressions, or more diverse task operators, remains an open issue. Third, because the candidate set is derived from syntactic analyses produced by an external parser, parsing errors or coverage limitations of the parser may directly affect the quality of the generated STL candidates. Finally, the current evaluation is limited to representative case studies rather than a large-scale benchmark.

These limitations suggest several directions for future work. An important next step is to handle sentences in which timing constraints are omitted or only implicitly specified, for example by incorporating default assumptions, context-dependent inference, or interactive clarification. Another direction is to expand the template library and the supported ambiguity types so that a wider range of natural-language task descriptions can be translated compositionally. It is also of interest to integrate the proposed candidate-generation framework with interactive disambiguation, downstream trajectory synthesis, and STL-based control design, so that candidate formulas can be selected, refined, and validated in a closed loop with the user and the control system. Finally, constructing a broader evaluation benchmark for ambiguity-aware NL-to-STL translation would be useful for comparing symbolic and learning-based approaches under a common protocol.


\bibliographystyle{ieeetr}
\bibliography{reference} 

\appendices
\section{Prompt and Evaluation Protocol for the ChatGPT Baseline}
\label{app:chatgpt_prompt}

We used ChatGPT (GPT-5.4 Thinking, accessed on \textit{2026/03/29}). 
Each input sentence was evaluated in a fresh chat session. The atomic propositions were defined as $\phi_A$, $\phi_B$, $\phi_C$, and $\phi_D$, indicating that the agent is in regions A, B, C, and D, respectively. The same prompt was used for S8--S12, with only the final sentence line replaced.

\noindent\textbf{Prompt used for each input sentence.}
\begin{quote}
You are given a navigation instruction and a mapping from region names to atomic propositions.\\ Use only the operators $\mathbf{F}_{[a,b]}$, $\mathbf{G}_{[a,b]}$, $\wedge$, $\vee$, and parentheses.\\ Do not output duplicates after simplification.\\ Atomic propositions: $\phi_A$: the agent is in region A; $\phi_B$: the agent is in region B; $\phi_C$: the agent is in region C; $\phi_D$: the agent is in region D.\\ Sentence: S8/S9/S10/S11/S12\\ Return only a numbered list of STL formulas.
\end{quote}

For evaluation, the raw ChatGPT outputs were compared with the manually identified reference interpretations after canonicalization. A raw output was marked correct if it matched a reference interpretation; otherwise, it was classified as one or more of the error types defined in Section~V-D.

\end{document}